\documentclass[11pt]{article}

\usepackage[margin=1in]{geometry}
\usepackage{amsmath,amssymb}
\usepackage{graphicx}
\usepackage{booktabs}
\usepackage{multirow}
\usepackage{hyperref}
\usepackage{caption}
\usepackage{subcaption}
\usepackage{cite}

\title{Noise Immunity in In-Context Tabular Learning: An Empirical Robustness Analysis of TabPFN's Attention Mechanisms}
\author{
\textbf{James Hu, Mahdi Ghelichi} \\
Model Development Innovation, Risk Management \\
TD Bank, Toronto, Canada \\
\texttt{\{james.hu, mahdi.ghelichi\}@td.com}
}
\date{}

\begin{document}
\maketitle

\begin{abstract}
Tabular foundation models (TFMs) such as TabPFN (Tabular Prior-Data Fitted Network) are designed to generalize across heterogeneous tabular datasets through in-context learning (ICL). They perform prediction in a single forward pass conditioned on labeled examples without dataset-specific parameter updates. This paradigm is particularly attractive in industrial domains (e.g., finance and healthcare) where tabular prediction is pervasive. Retraining a bespoke model for each new table can be costly or infeasible in these settings, while data quality issues such as irrelevant predictors, correlated feature groups, and label noise are common. 
In this paper, we provide strong empirical evidence that TabPFN is highly robust under these sub-optimal conditions. We study TabPFN and its attention mechanisms for binary classification problems with controlled synthetic perturbations that vary: (i) dataset width by injecting random uncorrelated features and by introducing nonlinearly correlated features, (ii) dataset size by increasing the number of training rows, and (iii) label quality by increasing the fraction of mislabeled targets. 
Beyond predictive performance, we analyze internal signals including attention concentration and attention-based feature ranking metrics. Across these parametric tests, TabPFN is remarkably resilient: ROC-AUC remains high, attention stays structured and sharp, and informative features are highly ranked by attention-based metrics. Qualitative visualizations with attention heatmaps, feature-token embeddings, and SHAP plots further support a consistent pattern across layers in which TabPFN increasingly concentrates on useful features while separating their signals from noise.
Together, these findings suggest that TabPFN is a robust TFM capable of maintaining both predictive performance and coherent internal behavior under various scenarios of data imperfections.
\end{abstract}

\section{Introduction}
Tabular prediction sits at the core of many industrial decision pipelines, including finance, healthcare, supply chain management, manufacturing, etc. where models support risk assessment, credit scoring, disease diagnosis, and operational triage at scale \cite{orion_msp, tabpfn_v1, grinsztajn2022whytree, van2024tabular}.
Despite the rise of unstructured data, tabular prediction is unique due to its inherent heterogeneity where datasets often feature a mix of variables with different meanings, varying scales, and complex non-linear relationships. Therefore, in these environments, nothing is perfectly clean: real tabular data commonly contain missing values, redundant and irrelevant predictors, correlated feature groups, outliers, and imperfect labels. A practical modeling approach must therefore be accurate and robust to low-quality artifacts that are unavoidable in the real world.
Historically, industry has relied heavily on classical tabular learners such as Random Forest, XGBoost, and CatBoost \cite{randomforest, xgboost, catboost}. These methods are strong in performance and relatively interpretable due to their robustness to feature scaling and their ability to handle missing values \cite{fernandez2014we, grinsztajn2022whytree, kursa2014robustness}. However, these classical models typically require dataset-specific training and hyperparameter tuning for every new task, which creates significant operational overhead, and their performance can depend strongly on the quality and quantity of labeled data available for that specific task \cite{kadra2021welltuned}.
This ``train-per-table'' workflow is often workable when the focus is a single dataset, but it becomes a bottleneck when the need to repeatedly build and maintain models across many heterogeneous tables with changing schemas and shifting data quality is imperative.

Recent progress in tabular foundation models (TFMs) offers a new alternative: instead of training a new model for each dataset, TFMs are pre-trained on a large and diverse collection of tables and then applied to unseen datasets without retraining. These models, often based on transformer architectures whose attention mechanisms allow them to focus on the most relevant features or interactions, may be able to ignore or infer low-quality or missing inputs by leveraging relationships elsewhere in the table \cite{attention}. These developments help shift the paradigm from per-dataset fitting to zero-shot in-context learning (ICL) \cite{ICL}. By conditioning on a labeled training set (the ``context'') during a single forward pass, TFMs approximate the posterior predictive distribution, without requiring gradient updates to the models' parameters. 
Tabular Prior-data Fitted Network (TabPFN) is a prominent example of this class. It is pre-trained once (offline) on massive synthetically generated datasets and then performs zero-shot inference on new tables using ICL both row-wise and column-wise across tabular cells \cite{tabpfn_v1, tabpfn_v2}.

However, while practitioners have strong intuitions about how tree-based learners behave under noisy or low-quality tabular inputs, there is an increasing need to build comparable understanding for TFMs: e.g. how do models like TabPFN behave when the table is wide, noisy, correlated, or partially mislabeled \cite{van2024tabular, labelnoise_tabpfn}?
In particular, robustness in industrial settings is not only about maintaining high prediction accuracy for certain datasets; it is also about whether internal signals indicate that the model consistently focuses on informative structure rather than spurious artifacts. 

In this paper, we study the robustness of TabPFN under controlled synthetic perturbations that directly mirror common industrial data issues. We vary (i) the number of features by injecting many irrelevant predictors, (ii) the presence of correlated feature groups, (iii) the number of training rows, and (iv) the proportion of label noise.
We evaluate predictive performance using ROC-AUC and complement it with internal diagnostics: attention concentration, attention-based feature ranking metrics, and qualitative visualizations of attention heatmaps, feature-token embedding structure, and SHAP value plots.
By linking robustness outcomes to interpretable internal signals, we aim to not only show that TabPFN remains stable in these regimes, but also understand how it does so, strengthening the case for further research and more confident deployment of TFMs in messy real-world environments. 

The organization of the paper is as follows. Section 2 discusses related work on TFMs and particularly TabPFN. In Section 3, we describe the experiment setups including the baseline study and parametric tests, and present the results to show the performance and robustness of the TabPFN model. Further implications from the results, limitations, and future work are discussed in Section 4. Section 5 concludes the paper.

\section{Related Work}
\paragraph{Tabular deep learning and the ICL paradigm.}
he evolution of deep learning for tabular data has progressed from specialized architectures like TabNet which uses sequential attention to prioritize the most important features \cite{tabnet} and FT-Transformer which also requires task-specific training \cite{gorishniy2021revisiting}, to foundation models capable of zero-shot inference. 
TabPFN pioneered this shift by framing tabular data problems as a meta-learning problem \cite{tabpfn_v1, tabpfn_v2}. 
The theoretical foundation of TabPFN is built on its ability to perform ICL via an approximation of Bayesian inference. Unlike traditional models that learn a specific rule and optimize model parameters for each new dataset, TabPFN is pre-trained on a collection of millions of synthetic datasets, referred to as a ``prior''. During this pre-training, the model learns the underlying statistical relationships across these tabular data in the prior. At inference time, given new tabular datasets, TabPFN can make predictions that favor explanations consistent with the pre-training prior. In the meantime, ICL enables TabPFN to perform these task adaptations by conditioning on examples rather than updating parameters in a forward pass.
The same works emphasize that transformer architectures can be adapted to tabular structure and that synthetic-data pre-training can expose the model to missingness, categorical variables, unimportant features, and outliers, leading to strong out-of-the-box performance on datasets within the studied regime \cite{tabpfn_v2}. 
Additional recent research has suggested that transformers with ICL may internally implement algorithms similar to iterative gradient descent, allowing them to serve as optimizers for new tables during the forward pass \cite{von2023transformers}.

\paragraph{Handling heterogeneity via tokenization and randomized attribute tokens.}
A critical challenge for tabular foundation models is heterogeneity: datasets differ in dimensionality and in the meaning of attributes, and attribute semantics may be unavailable (e.g., due to privacy or lack of descriptive names). A detailed analysis of TabPFN v2 argues that the model can handle heterogeneity using randomized attribute tokens resampled at inference time, and that the model can infer attribute relationships through ICL despite randomness in tokenization \cite{closerlook_tabpfnv2}. The same study also shows how TabPFN v2 can be repurposed as a feature extractor and discusses post-hoc divide-and-conquer strategies to mitigate limitations in high-dimensional, many-class, and large-scale regimes without retraining.

\paragraph{Scalability constraints and architectural responses.}
While TabPFN with its ICL mechanism is compelling, attention that is both row-wise and feature-wise dense introduces unfavorable scaling with table width, as the computational cost is quadratic with respect to the number of samples and the number of features \cite{tabpfn_v2}. 
Orion-MSP attempts to solve the limited context-window with multi-scale feature processing and block-sparse attention \cite{orion_msp}.
TabDPT, trained on real-world data with self-supervised learning, increases the scalability from its row-based token and retrieval-based context selection that uses top most similar rows as context \cite{tabdpt}. Other approaches, discussed in \cite{closerlook_tabpfnv2}, utilize divide-and-conquer strategies or test-time scaling to manage high-dimensional data \cite{wei2022chain, muennighoff2025s1}.
Our work complements these structural discussions by focusing on the qualitative integrity of the already developed TabPFN models.

\paragraph{Positioning of this work and robustness evaluation beyond predictive metrics.}
Most robustness discussions for tabular datasets focus on performance metrics under dataset shifts or structural changes \cite{tabarena, tanna2026exploring, tabpfn_v2}, while fewer studies connect robustness to internal model signals. 
Building on this perspective, our study explicitly combines predictive evaluation using ROC-AUC with internal attention-based metrics, representation-level diagnostics, and interpretability capabilities such as SHAP-based feature importance introduced in \cite{shap} to probe whether robustness corresponds to stable internal prioritization of informative features under perturbations.
Relative to the above literature, our contribution is a targeted robustness analysis that 
(1) uses controlled synthetic perturbations to isolate factors such as feature noise, correlation, sample size, and label noise and
(2) links performance stability to interpretable internal signals. 
This can aid benchmark-level claims by making the mechanism-explanation relationship more explicit under controlled stress tests.

\section{Robustness Experiments}
In this section, we describe our experiment design to test robustness of the TabPFN model and present the performance results. 
First we conduct a baseline study on a fixed synthetic binary classification dataset with 1500 training rows, 2 informative features and 6 random features. We aim to show and discuss what happens inside TabPFN with visualizations when it learns from a dataset that has both informative and random features.
To investigate how the quality of datasets affects performance of TabPFN, we further define four parametric tests, keeping the data generation controlled and varying one factor at a time unless stated otherwise:
\begin{enumerate}
 \item{Impact of feature dimension}:
 \begin{enumerate}
  \item{Random features}: fixing number of rows at 1500 and number of informative features at 2, we increase total features from 4 to 512 by adding random, uncorrelated features.
  \item{Correlated features}: fixing number of total features at 512 and number of informative features at 2, we increase number of nonlinearly correlated features from 1 to 128.
 \end{enumerate}
 \item{Impact of sample size}: fixing number of number of total features at 8 with 2 informative features and 6 random features, we increase number of rows from 1500 to 12000.
 \item{Impact of label noise}: fixing number of rows at 6000, total number of features at 16 with 2 informative features and 14 random features, we vary the fraction of label noise from 0 to 0.35 by randomly mislabeling the targets.
\end{enumerate}
The purpose of the setups is to separate distinct sources of difficulty that frequently arise in tabular modeling in the real world. In practice, performance degradation can occur because a table becomes wider and contains more irrelevant variables, because correlated predictors compete with genuinely informative ones, because the amount of available data changes, or because the supervision itself is noisy. By varying each source of difficulty in isolation, we aim to determine whether TabPFN remains stable under each variation and whether any change in predictive performance is accompanied by a change in the model's internal attention behavior.

\paragraph{Evaluation Metrics.}
We evaluate ROC-AUC on a held-out test dataset of the same size as the primary predictive performance metric. For attention-based metrics, we quantify how TabPFN allocates attention across feature tokens by computing:
\begin{enumerate}
  \item{Informative attention proportion}:
  the fraction of total attention assigned to the 2 informative features.
  \item{Attention ratio}:
  mean of attention on informative features divided by mean of attention on all other features.
  \item{Ranking quality}:
  mean rank of informative features and the proportion of the informative features in top-2 attended features.
  \item{Attention concentration}:
  $KL_1 = $ KL divergence of the attention distribution relative to a uniform distribution (higher values indicate more concentrated and structured attention); $KL_2 = $ KL divergence of the attention distribution relative a distribution where informative features are weighted equally and all other random features are assigned with small weights $w_{\epsilon}=10^{-6}$ (lower values indicate that attention is more concentrated on the informative features and is distributed more evenly across them.)
\end{enumerate}

Qualitative analyses are included in both the baseline study and parametric tests with visualizations of
(1) attention heatmaps across layers to assess distribution of attention inside the model between all features and the label;
(2) feature token embedding projections across layers to assess separability between informative and other features;
(3) SHAP global importance and per-sample attributions to compare post-hoc feature importance with attention patterns.
Note that in these synthetic datasets, the indices of features are always in the order of informative, correlated if any, and random features, e.g., the first 2 columns represent 2 informative features.

\subsection{Baseline Analysis}
For the baseline dataset, TabPFN gives an accurate prediction with ROC-AUC = 0.974 on the test data. We examine what happens inside TabPFN and show that TabPFN can focus on useful information from the dataset with the presence of random features.
\subsubsection{Attention Concentration Across Layers.}
\begin{figure}[h]
\centering
\includegraphics[width=0.48\textwidth]{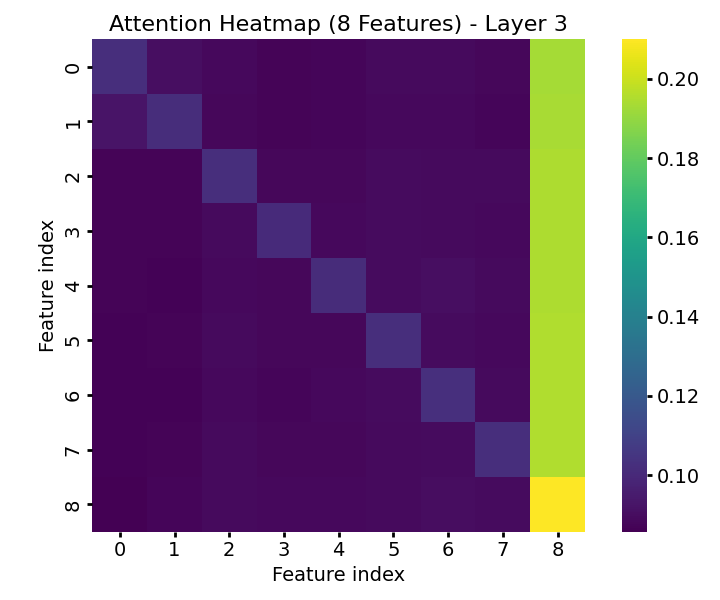}
\includegraphics[width=0.48\textwidth]{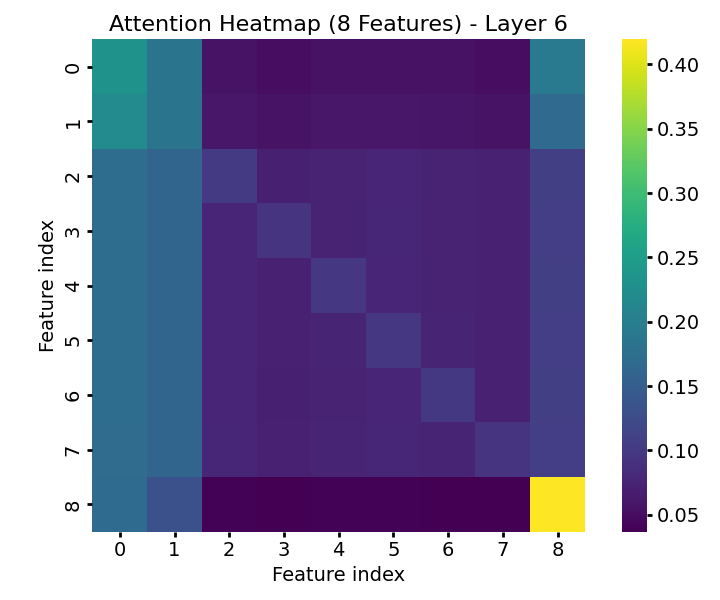}
\includegraphics[width=0.48\textwidth]{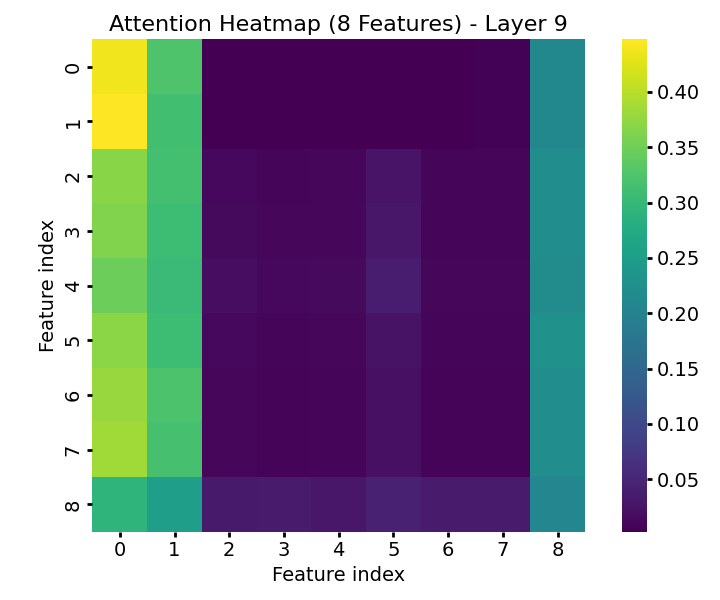}
\includegraphics[width=0.48\textwidth]{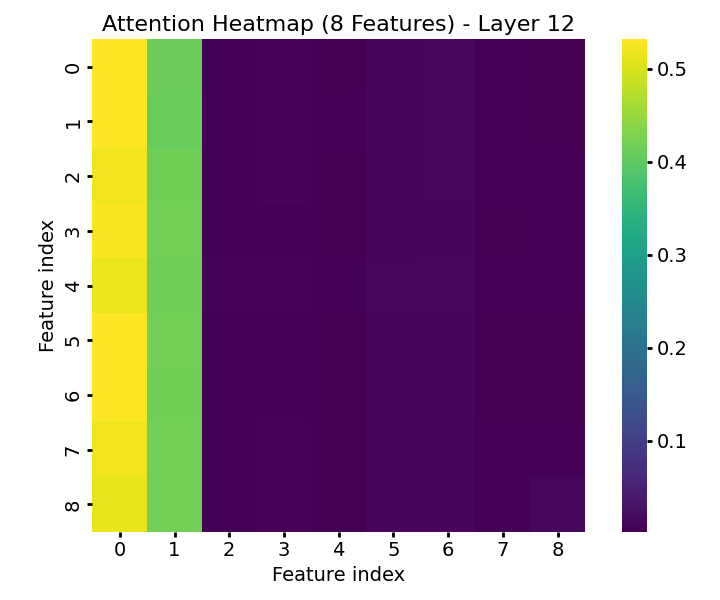}
\caption{Feature-wise attention weights heatmaps in layer \{3, 6, 9, 12\} of TabPFN for the baseline case. In each plot, the first 2 indices represent informative features and last index is the target label.}
\label{fig:baseline_attn_heatmap}
\end{figure}

To visualize the attention distribution among features, we extract the attention weights on all features and the label token. Fig.~\ref{fig:baseline_attn_heatmap} shows the heatmaps of attention weights averaged over rows and heads. Each column in a heatmap represents the attention received by one feature from all feature and label tokens. The first 2 feature indices correspond to the informative features and the rest are random features. The last column and row are the attention to and from the label respectively.
Attention heatmaps show a temporal learning pattern across layers. In the early layer (layer 3) attention is emphasized on the label token to capture task information. In middle layers (layer 6 and layer 9), attention is spread across features to discover feature interactions and then become concentrated onto the 2 informative features. In the last transformer layer (layer 12), attention is focused on the 2 informative features. This provides a mechanistic hint for how the prediction remains strong even when there are random features in the data.

\subsubsection{Feature Token Embedding Separation.}
\begin{figure}[h]
\centering
\includegraphics[width=0.48\textwidth]{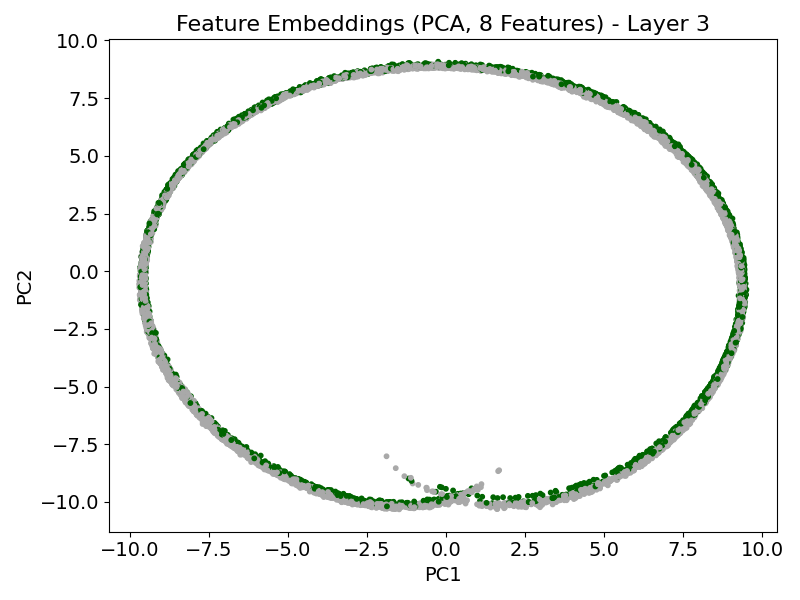}
\includegraphics[width=0.48\textwidth]{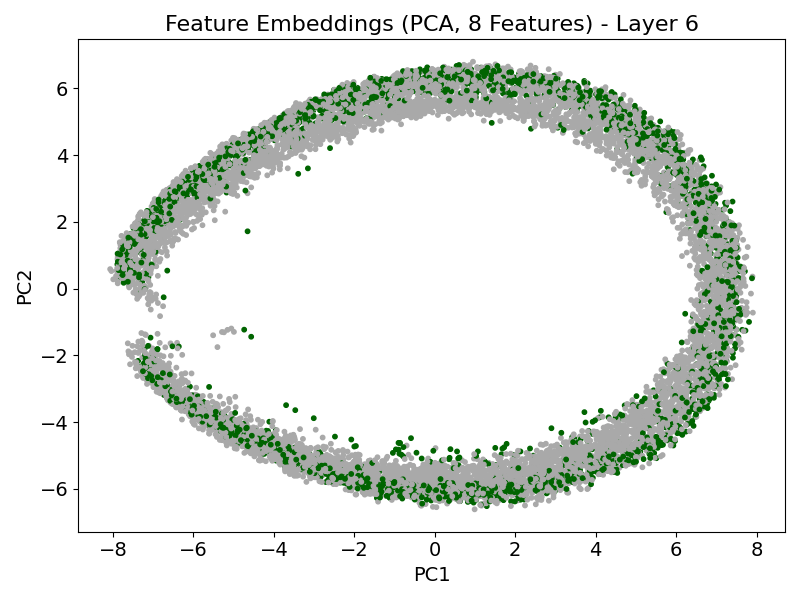}
\includegraphics[width=0.48\textwidth]{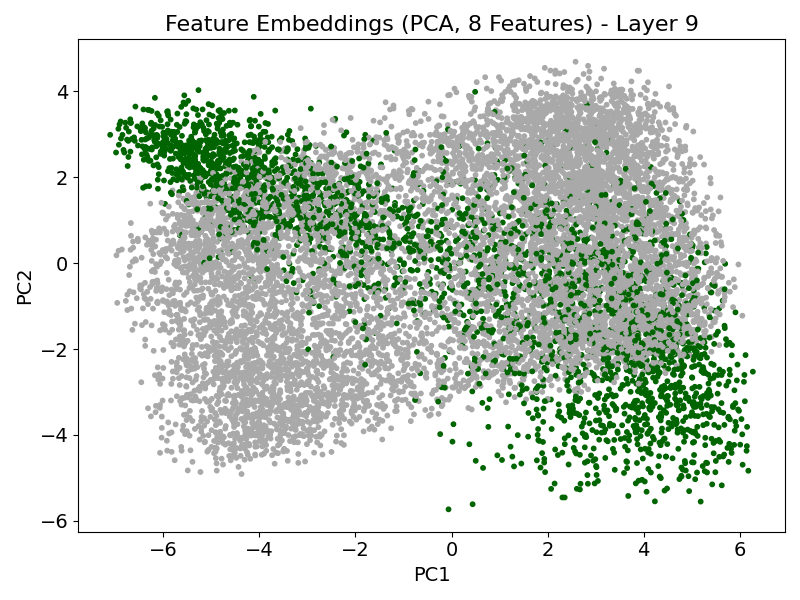}
\includegraphics[width=0.48\textwidth]{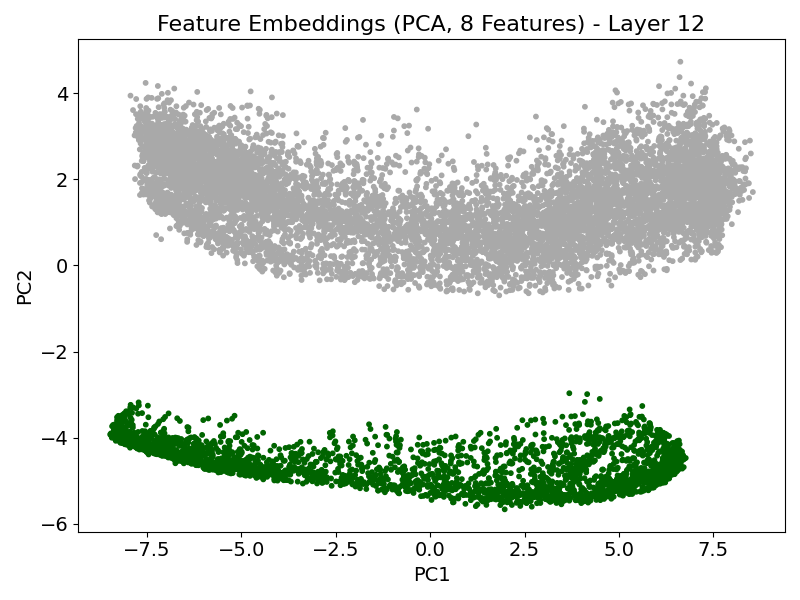}
\caption{Feature-token embeddings in layer \{3, 6, 9, 12\} of TabPFN for the baseline case. In each plot, the green points represent informative features and gray points are random features.}
\label{fig:baseline_embed}
\end{figure}

To provide a geometric interpretation of the attention pattern, we visualize the evolution of feature token embeddings across transformer layers.
We project these high-dimensional feature embeddings into a 2D space using Principal Component Analysis (PCA), where informative features are colored in green and random features are colored in gray in Fig.~\ref{fig:baseline_embed}. It is observed that in early layers (layer 3), the informative and random features are entangled and indistinguishable. In middle layers (layer 6 and layer 9), these representations begin to shift and become sharper. In the last transformer layer (layer 12), the embeddings are linearly separable, where the informative features form distinct clusters isolated from the random features.
This suggests that TabPFN internally clusters and distinguishes useful from non-informative features, and can provide a consistent conclusion with observations from the attention heatmaps.

\subsubsection{SHAP Alignment With Attention.}
\begin{figure}[h]
\centering
\includegraphics[width=0.41\textwidth]{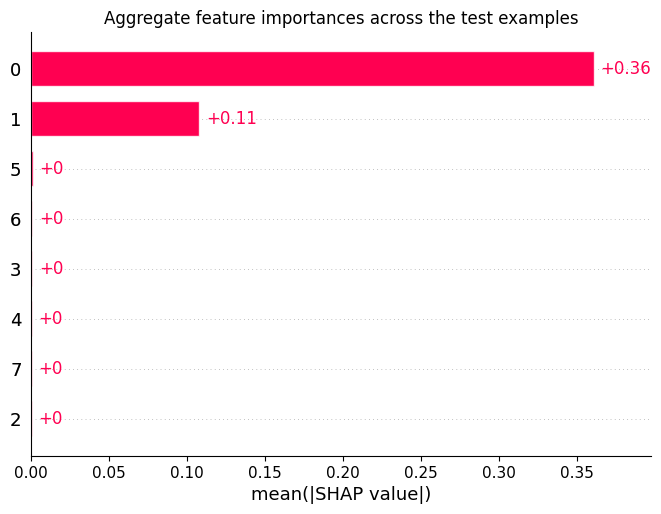}
\includegraphics[width=0.56\textwidth]{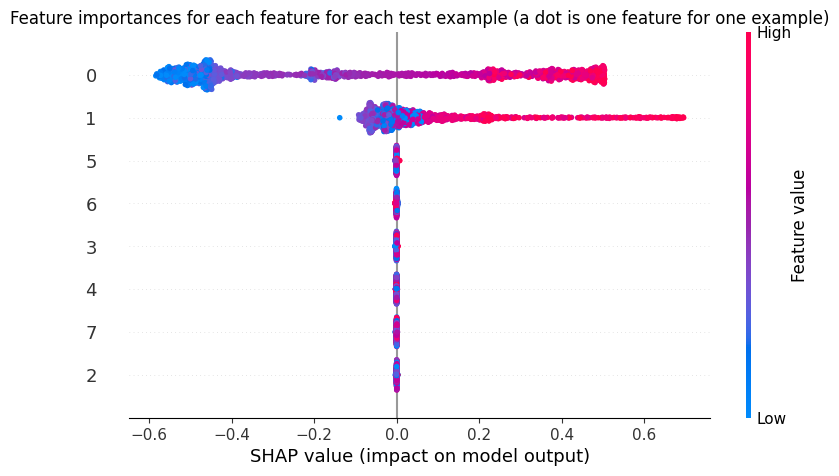}
\caption{SHAP plots of TabPFN for the baseline case. First two indices represent informative features, and the rest are random features. Left: aggregate SHAP values across samples for all features; right: per-sample SHAP values distribution for all features.}
\label{fig:baseline_shap}
\end{figure}

SHAP is commonly used to quantify how much each feature contributes to a specific prediction \cite{shap}.
To analyze TabPFN's feature attributions, we calculate and plot SHAP values that explain the model's output for the first class in our binary classification data. While SHAP values are model dependent, they allow us to visualize which features TabPFN prioritizes and determine the model's internal consistency.
Fig.~\ref{fig:baseline_shap} provides evidence that TabPFN prioritizes useful signals. Specifically, the 2 informative features, feature 0 and feature 1, exhibit dominant global aggregate SHAP values and are the main drivers of model predictions. The remaining random features yield negligible SHAP values, confirming TabPFN has filtered out this non-informative noise. We also note that feature 0 has higher importance score than feature 1, and it is consistent with observations from heatmaps that feature 0 receives more attention than feature 1.
In addition, per-sample SHAP values also show larger, varying attributions for informative features whereas random features have minimal attributions. 
These SHAP importance results align with and support previous conclusions from attention heatmaps and feature token embeddings: TabPFN learns to rely primarily on informative features and downweights random features.

\subsection{Parametric Test: Robustness to Increasing the Number of Features}
We first study robustness with respect to feature dimensionality. This family of tests is intended to distinguish two different scenarios. In the first scenario, the model is exposed to a growing number of irrelevant predictors that are random and should ideally be ignored. In the second scenario, the additional predictors are nonlinearly correlated with the underlying signal and may therefore act as plausible but redundant alternatives to the informative features. These two cases test different aspects of robustness: resistance to feature clutter versus stability under redundancy and confounding structure. 
\subsubsection{Random Features}
\begin{figure}[h]
\centering
\begin{subfigure}[b]{0.48\textwidth}
\centering
\includegraphics[width=\linewidth]{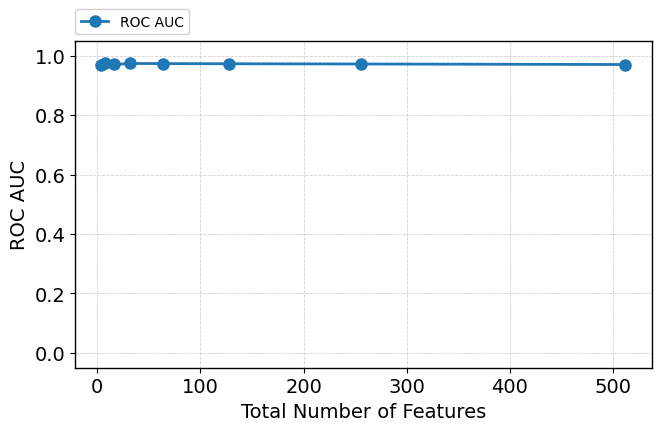}
\end{subfigure}
\begin{subfigure}[b]{0.48\textwidth}
\centering
\includegraphics[width=\linewidth]{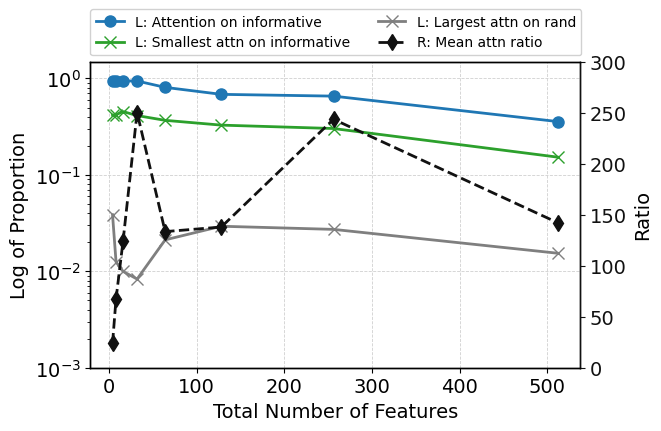}
\end{subfigure}
\par
\vspace{0.5\baselineskip}
\begin{subfigure}[b]{0.48\textwidth}
\centering
\includegraphics[width=\linewidth]{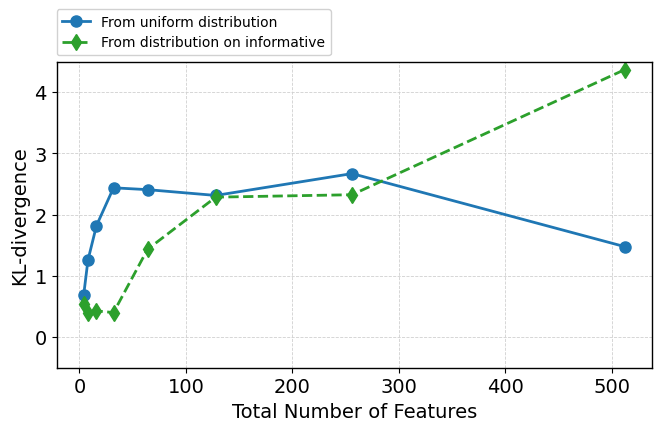}
\end{subfigure}
\begin{subfigure}[b]{0.48\textwidth}
\centering
\includegraphics[width=\linewidth]{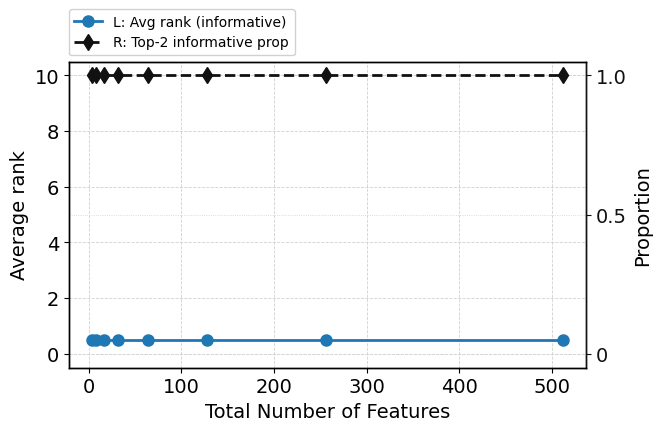}
\end{subfigure}
\caption{Performance and attention metrics of TabPFN with respect to increasing number of random features. Top left: ROC-AUC; top right: attention proportions and attention ratio; bottom left: KL-divergence with respect to different distributions; bottom right: average attention rank of informative features and their proportion in top-2 attended features.}
\label{fig:parametric_randomfea}
\end{figure}

In the first setting, we increase the total number of features from 4 to 512 by adding uncorrelated random features while keeping the number of informative features fixed at 2.
In Fig.~\ref{fig:parametric_randomfea}, we observe that the ROC-AUC remains high and stable as the number of random features increases, indicating that widening the table with irrelevant predictors does not degrade predictive performance in this test. 
At the same time, the attention distribution remains non-uniform, with $KL_1$ (KL divergence relative to the uniform distribution) consistently greater than 0.2. This indicates that TabPFN's attention is sharp and does not simply spread attention uniformly across all available columns as feature dimension grows. 
Furthermore, $KL_2$ (KL divergence relative to the distribution with equal weights on informative features, and all others small weights) increases as the number of random feature grows, and is also much larger than 0. We note that this does not necessarily mean weights are not concentrated on informative features. Rather, it hints that TabPFN can prioritize one of the informative features over another, deviating from the distribution where weights are equally distributed between 2 informative features. In addition, the large $KL_2$ value can result from some random features that get small portions of weights larger than their defined weights $w_\epsilon = 10^{-6}$.
While the informative attention proportion may decrease as total feature count grows so that the denominator in the proportion increases, the model does not shift substantial attention mass to random features, and informative-vs-noise attention ratios remain high. This conclusion is further supported by the ranking metrics which show informative features remain top-ranked (mean rank near 0.5; top-2 proportion equals 1 in the reported tests). 

\begin{figure}[h]
\centering
\includegraphics[width=0.48\textwidth]{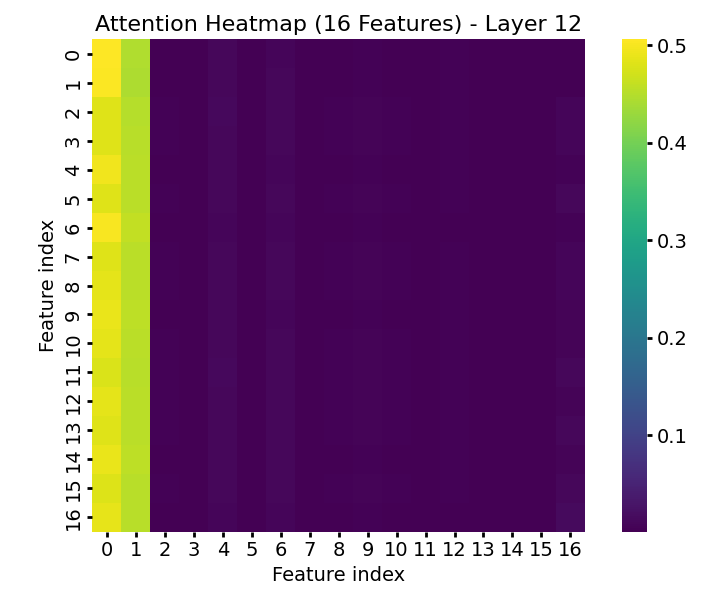}
\includegraphics[width=0.48\textwidth]{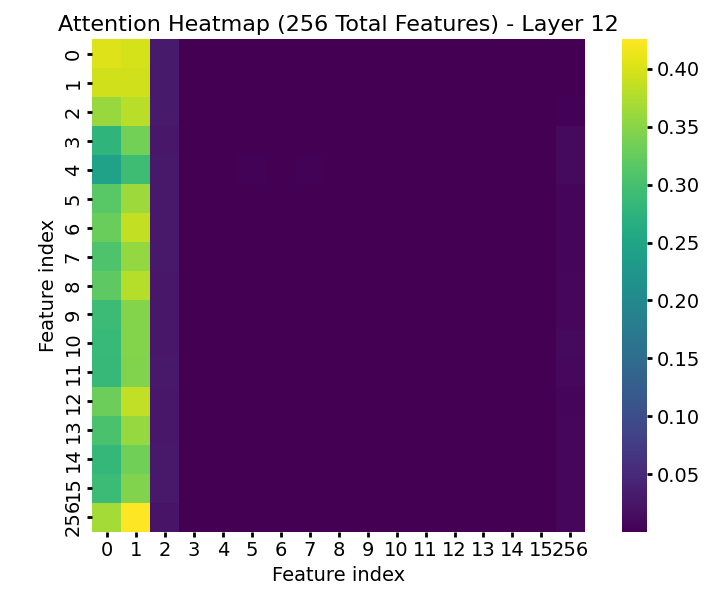}
\caption{Feature-wise attention weights heatmaps in layer 12 of TabPFN. In each plot, the first 2 indices represent informative features and last index is the target label. Left: total number of features = 16; right: total number of features = 256, the first 16 features and the target label are shown.}
\label{fig:bytotalfea_attn_heatmap}
\end{figure}

To further visualize attention concentration on the 2 informative features, Fig.~\ref{fig:bytotalfea_attn_heatmap} shows attention heatmaps for 2 cases when total numbers of features are 16 and 256. In the plot when total number of features is 256, only the first 16 features and the target label are shown for comparison and readability. We can see that most of the attention is received by the first 2 informative features, although they are not equally weighted. These support the ranking metrics and TabPFN is certain in prioritizing informative signals.

\subsubsection{Correlated Features}
\begin{figure}[h]
\centering
\begin{subfigure}[b]{0.48\textwidth}
\centering
\includegraphics[width=\linewidth]{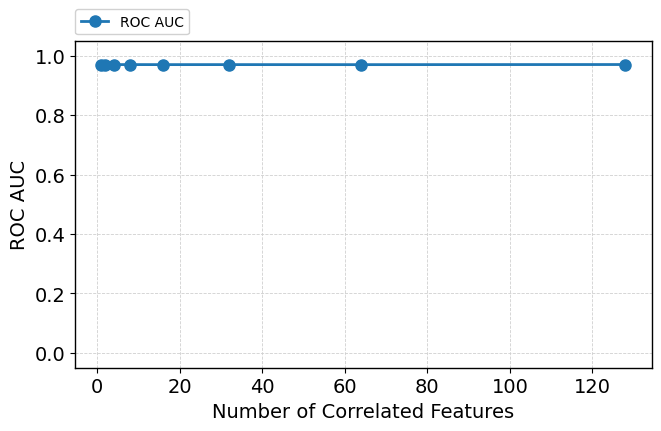}
\end{subfigure}
\begin{subfigure}[b]{0.48\textwidth}
\centering
\includegraphics[width=\linewidth]{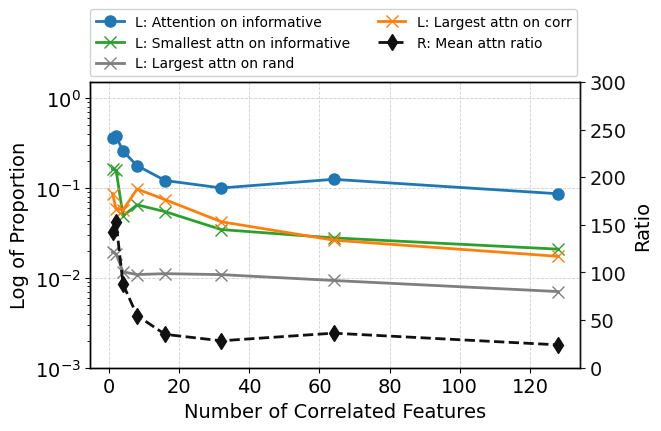}
\end{subfigure}
\par
\vspace{0.5\baselineskip}
\begin{subfigure}[b]{0.48\textwidth}
\centering
\includegraphics[width=\linewidth]{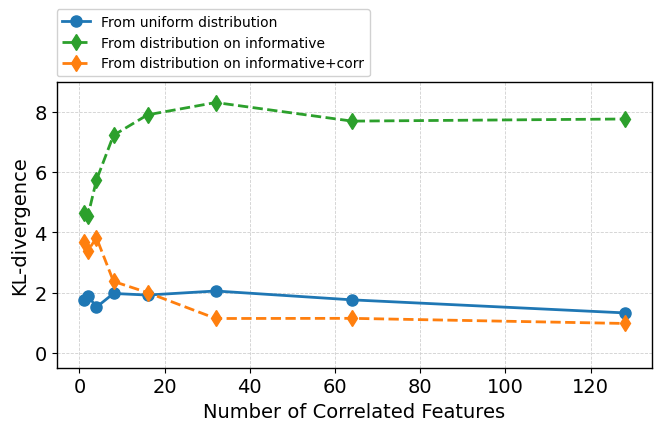}
\end{subfigure}
\begin{subfigure}[b]{0.48\textwidth}
\centering
\includegraphics[width=\linewidth]{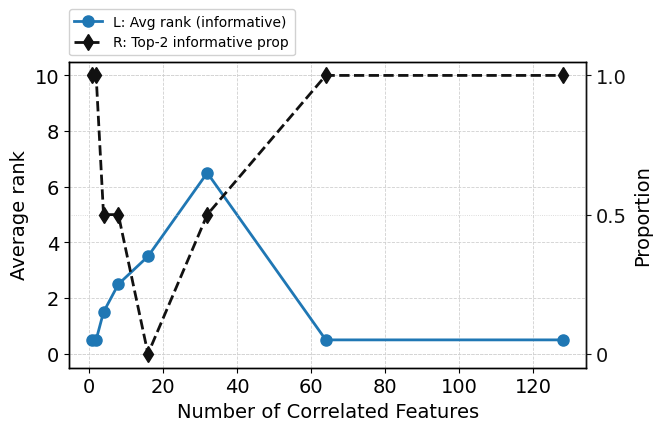}
\end{subfigure}
\caption{Performance and attention metrics of TabPFN with respect to increasing number of nonlinearly correlated features. Top left
: ROC-AUC; top right: attention proportions and attention ratio; bottom left: KL-divergence with respect to different distributions; bottom right: average attention rank of informative features and their proportion in top-2 attended features.}
\label{fig:parametric_corrfea}
\end{figure}

In the second setting, we investigate the model's behavior under feature redundancy. While random features are easy to separate from informative features, nonlinearly correlated features present a more sophisticated challenge for TabPFN's attention mechanism. Here we increase the number of correlated features derived from nonlinear transformations of the 2 informative features from 1 to 128, while keeping the total number of features fixed at 512.
When correlated features are introduced, ROC-AUC again remains high without any performance drop as in Fig.~\ref{fig:parametric_corrfea}.
$KL_1$ remains positive and is much larger than 0, indicating structured attention. For this setting, we also compute $KL_3$ defined as the KL divergence relative to a distribution where informative and correlated features are equally weighted, and all other random features have small weights. This way, we can compare $KL_3$ with $KL_2$ to see the effects of adding correlated features. High $KL_2$ value shows weight distributions are not exclusively and equally on 2 informative features. $KL_3$ supports the observation as its values are smaller than both $KL_1$ and $KL_2$ when the number of correlated features is large. This indicates that correlated features compete with informative features for attention.  
In the proportion plot, it is similarly observed that rather than attention remaining strictly focused on the original 2 informative features, the model begins to distribute weights across the correlated features as the proportion of attention on informative features declines in the beginning and we observe occasional cases where informative features are not among the top-2 attended features. We note that, in some cases, informative features exhibit only slightly larger average attention-rank values, suggesting that they are not substantially deprioritized relative to many other features.
Although the attention ratio of informative features over all other features also decreases initially, its value remains at a high level. This suggests that TabPFN recognizes the shared mutual information between informative and correlated features. 

\begin{figure}[h]
\centering
\includegraphics[width=0.44\textwidth]{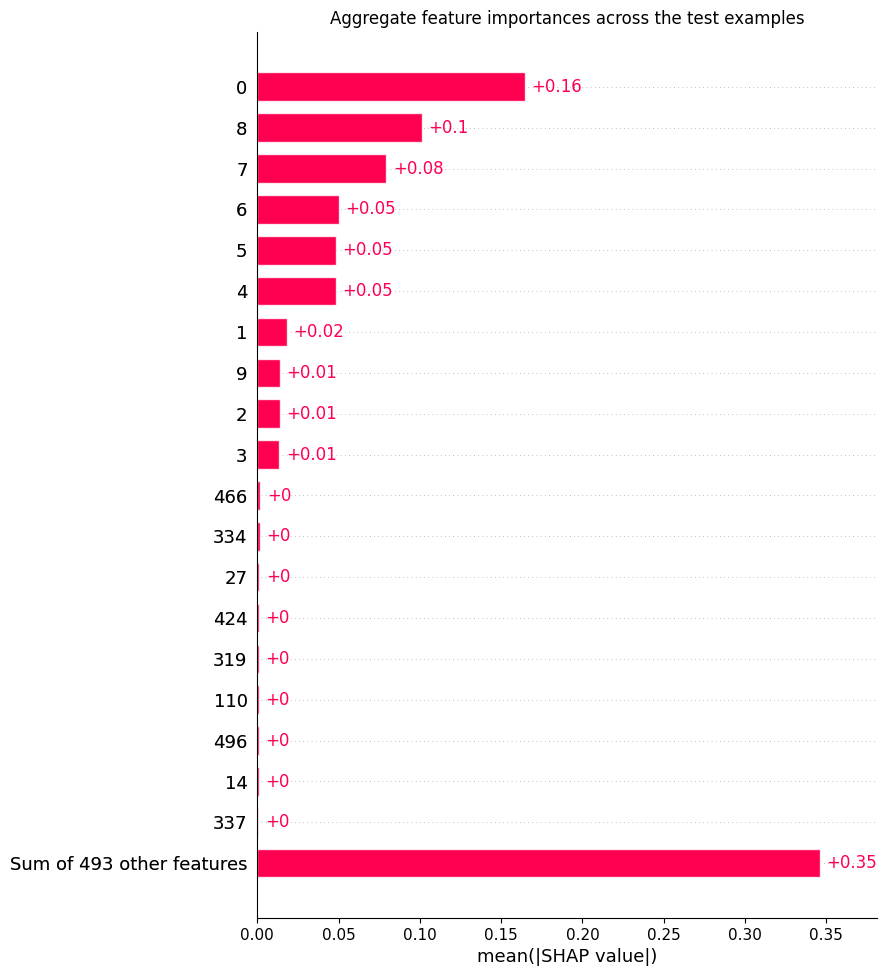}
\includegraphics[width=0.42\textwidth]{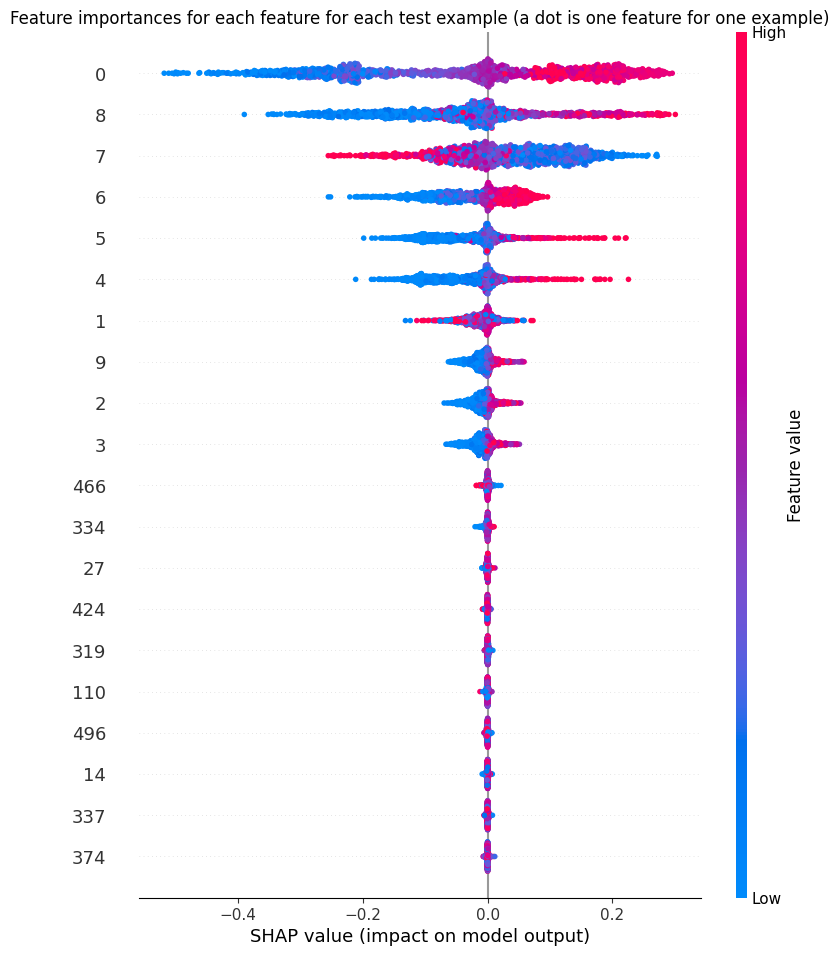}
\caption{SHAP plots of TabPFN when the number of correlated features = 8. First two indices represent informative features, subsequent 8 indices are correlated features, and the rest are random features. Left: aggregate SHAP values across samples for all features; right: per-sample SHAP values distribution for top-20 features.}
\label{fig:corr_shap}
\end{figure}

SHAP plots when the number of correlated features is 8 are shown in Fig.~\ref{fig:corr_shap} where indices 0 and 1 represent informative features, indices 2 to 9 are correlated features, and the rest are random features. Since we see attention does focus on the correlated features, their SHAP values are higher as expected. Feature 0 is still the most important feature whereas feature 1 drops out of top-2 which is consistent with our ranking metrics shown in Fig.~\ref{fig:parametric_corrfea}.
Notably, this ``crowded'' competition for attention from informative and correlated features does not translate into degraded predictive performance in this test yet may cause explainability issues for TabPFN. Further discussion is provided in Section 4.

\subsection{Parametric Test: Robustness to Increasing the Number of Training Rows}
\begin{figure}[h]
\centering
\begin{subfigure}[b]{0.48\textwidth}
\centering
\includegraphics[width=\linewidth]{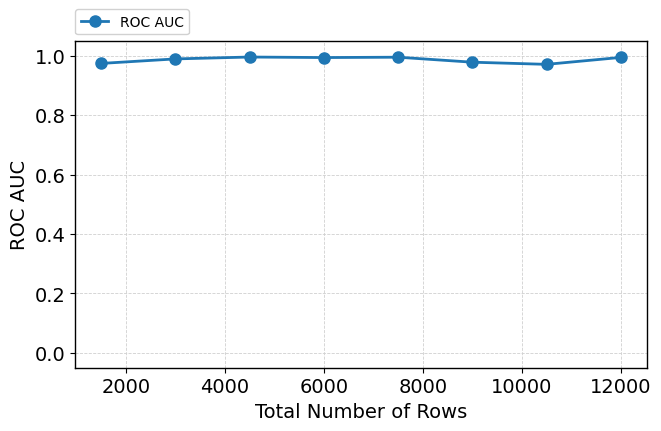}
\end{subfigure}
\begin{subfigure}[b]{0.48\textwidth}
\centering
\includegraphics[width=\linewidth]{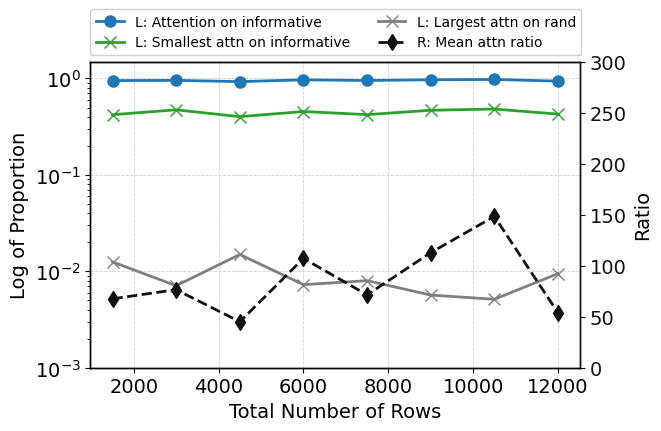}
\end{subfigure}
\par
\vspace{0.5\baselineskip}
\begin{subfigure}[b]{0.48\textwidth}
\centering
\includegraphics[width=\linewidth]{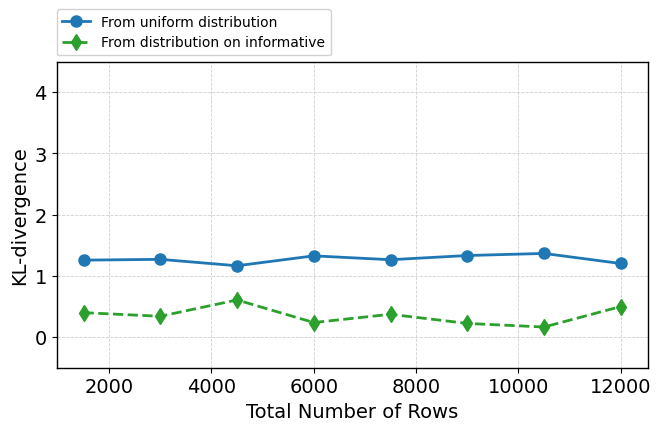}
\end{subfigure}
\begin{subfigure}[b]{0.48\textwidth}
\centering
\includegraphics[width=\linewidth]{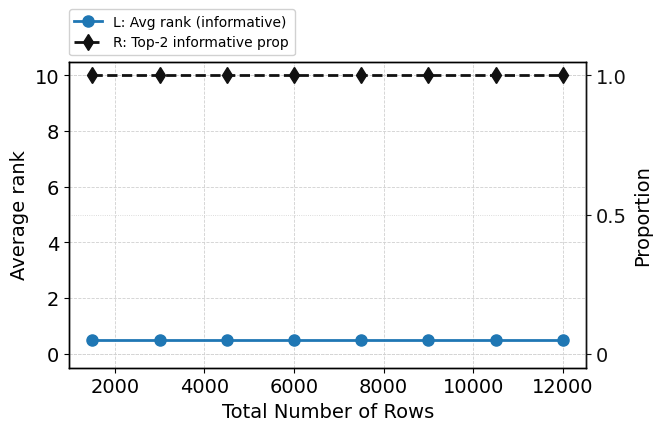}
\end{subfigure}
\caption{Performance and attention metrics of TabPFN with respect to increasing the number of rows. Top left: ROC-AUC; top right: attention proportions and attention ratio; bottom left: KL-divergence with respect to different distributions; bottom right: average attention rank of informative features and their proportion in top-2 attended features.}
\label{fig:parametric_rows}
\end{figure}

To test sample-wise scalability, we expand the training set from 1500 to 12000 rows to test how increasing sample size affects TabPFN's robustness. Because TabPFN's complexity is quadratic with respect to sample size \cite{tabpfn_v2}, this test allows us to examine whether increasing the number of rows makes the model less certain in its attention allocation and if the model remains robust in performance. As the number of training rows grows, we find that ROC-AUC remains stable at high levels and attention remains highly structured where $KL_1$ is always greater than 1 from Fig.~\ref{fig:parametric_rows}. Low $KL_2$ further suggests that most weights are distributed on 2 informative features. Since $KL_2$ is still noticeably greater than 0 for most cases, the weights are not equally distributed between 2 informative features.
Ranking metrics continue to prioritize informative features, and attention proportion and its ratio over other features stay high, suggesting that the model is certain about which features are driving the label and adding data does not weaken the model's focus on informative features.

\begin{figure}[h]
\centering
\includegraphics[width=0.48\textwidth]{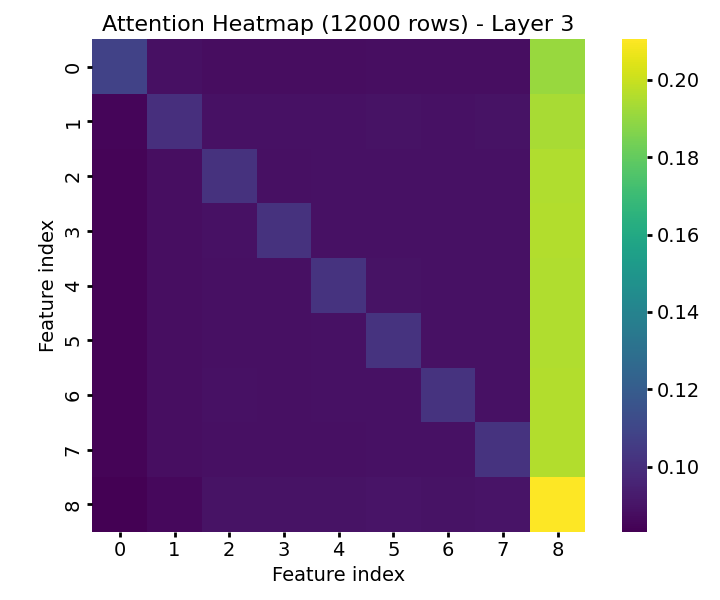}
\includegraphics[width=0.48\textwidth]{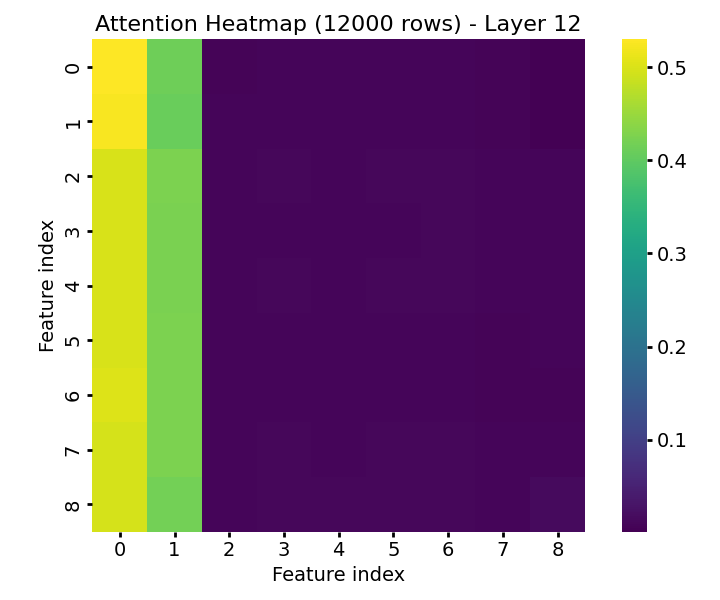}
\includegraphics[width=0.48\textwidth]{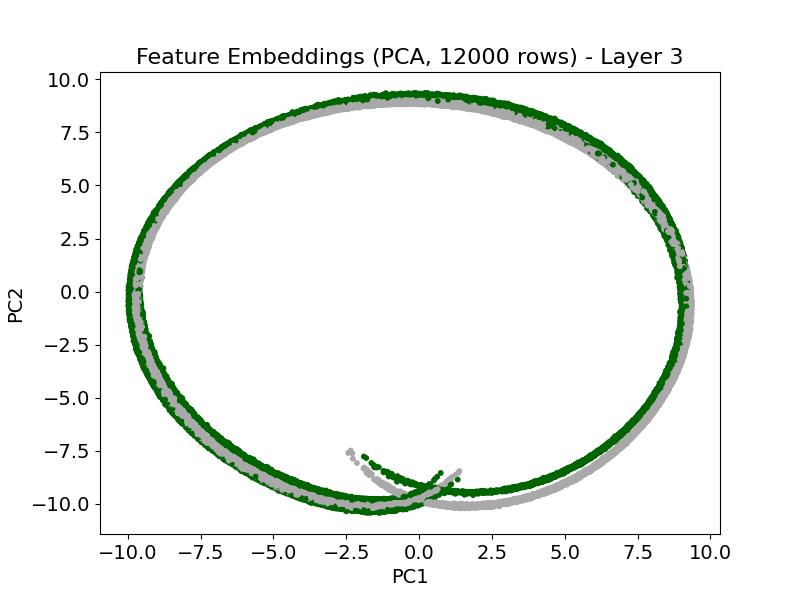}
\includegraphics[width=0.48\textwidth]{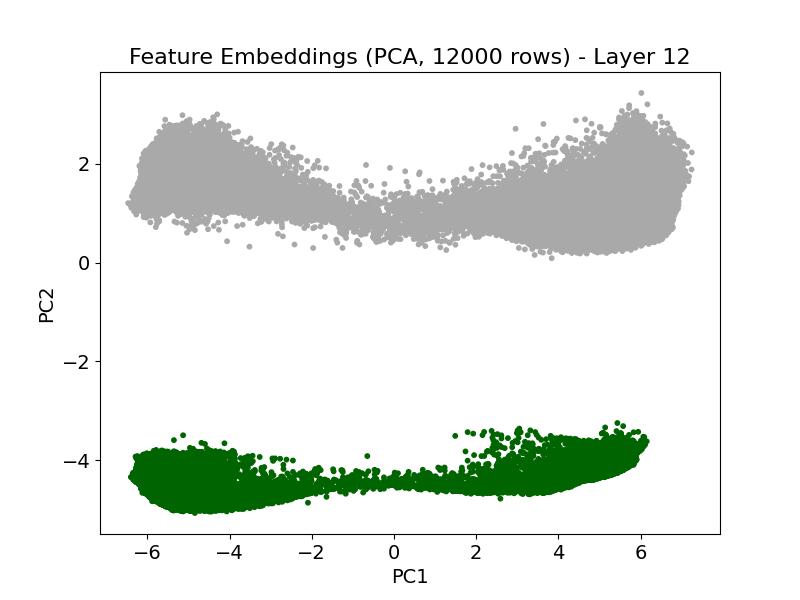}
\caption{Feature-wise attention weights heatmaps and feature-token embeddings in layer \{3, 12\} of TabPFN when number of rows = 12000. In each attention heatmap plot, the first 2 indices represent informative features and last index is the target label. In each embeddings plot, the green points represent informative features and gray points are random features. Top left: attention heatmap in layer 3; top right: attention heatmap in layer 12; bottom left: feature embedding plot in layer 3; bottom right: feature embedding plot in layer 12.}
\label{fig:byrows_attn_heatmap_embed}
\end{figure}

To strengthen the conclusion and compare with our baseline case when the number of rows = 1500, we show attention heatmaps and feature-token embeddings in layer 3 and layer 12 when the number of rows = 12000 in Fig.~\ref{fig:byrows_attn_heatmap_embed}. Similar behaviors are seen as in the baseline case. In early layer 3, attention is focused on the label and all features are mixed together. In the last transformer layer 12, 2 informative features are mostly attended and their embeddings become sharply separated from random features.

\subsection{Parametric Test: Robustness to Increasing the Proportion of Label Noise}
\begin{figure}[h]
\centering
\begin{subfigure}[b]{0.48\textwidth}
\centering
\includegraphics[width=\linewidth]{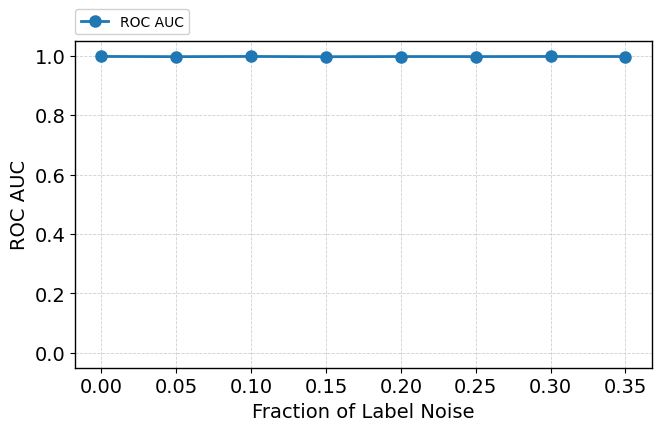}
\end{subfigure}
\begin{subfigure}[b]{0.48\textwidth}
\centering
\includegraphics[width=\linewidth]{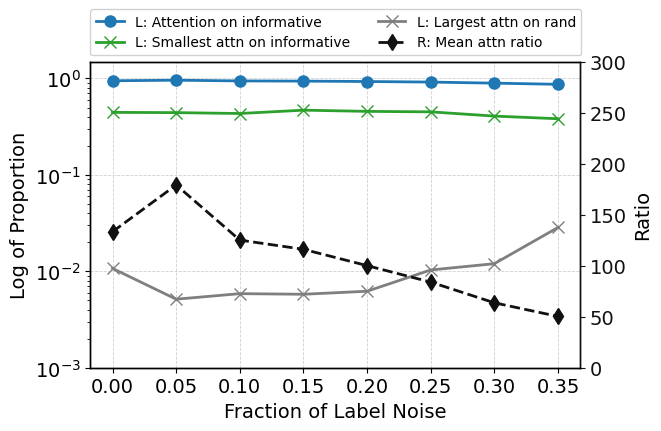}
\end{subfigure}
\par
\vspace{0.5\baselineskip}
\begin{subfigure}[b]{0.48\textwidth}
\centering
\includegraphics[width=\linewidth]{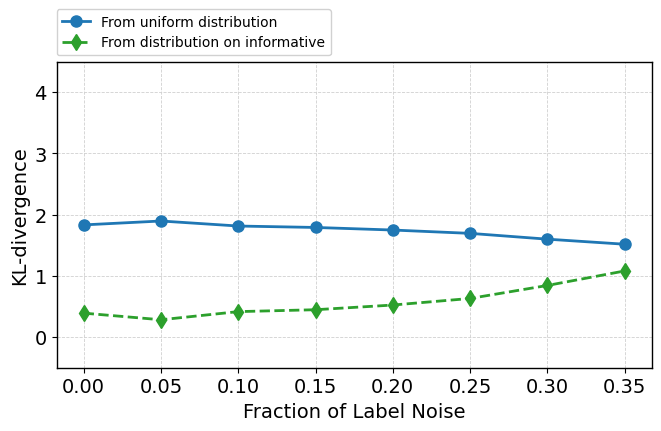}
\end{subfigure}
\begin{subfigure}[b]{0.48\textwidth}
\centering
\includegraphics[width=\linewidth]{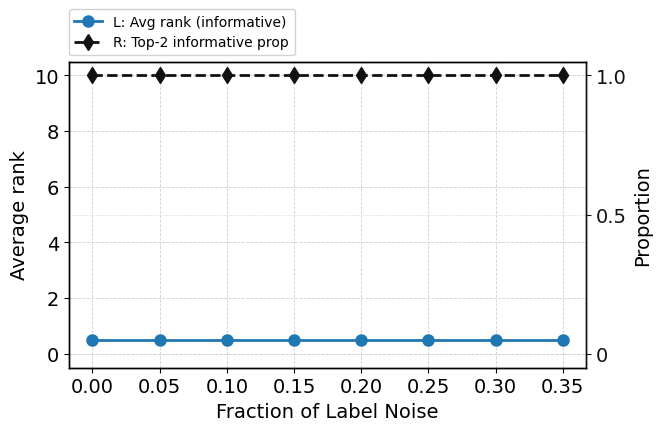}
\end{subfigure}
\caption{Performance and attention metrics of TabPFN with respect to increasing proportion of label noise. Top left: ROC-AUC; top right: attention proportions and attention ratio; bottom left: KL-divergence with respect to different distributions; bottom right: average attention rank of informative features and their proportion in top-2 attended features.}
\label{fig:parametric_labelnoise}
\end{figure}

Here we evaluate TabPFN's resilience to random mislabeling of the target. As in the real world, labels are frequently misrecorded, a robust model should avoid overfitting or mistakenly learning to this noise. For this test, we increase the proportion of the label noise from none to 35\% by randomly flipping the binary classification label. 
As shown in Fig.~\ref{fig:parametric_labelnoise}, ROC-AUC does not degrade. Attention structure persists as $KL_1$ is larger than 1 and most weights are concentrated on 2 informative features since $KL_2$ value is relatively small in all cases. Similar to the previous test with increasing number of rows, informative features stay top-ranked and attention-based ratios remain high. 

\begin{figure}[h]
\centering
\includegraphics[width=0.48\textwidth]{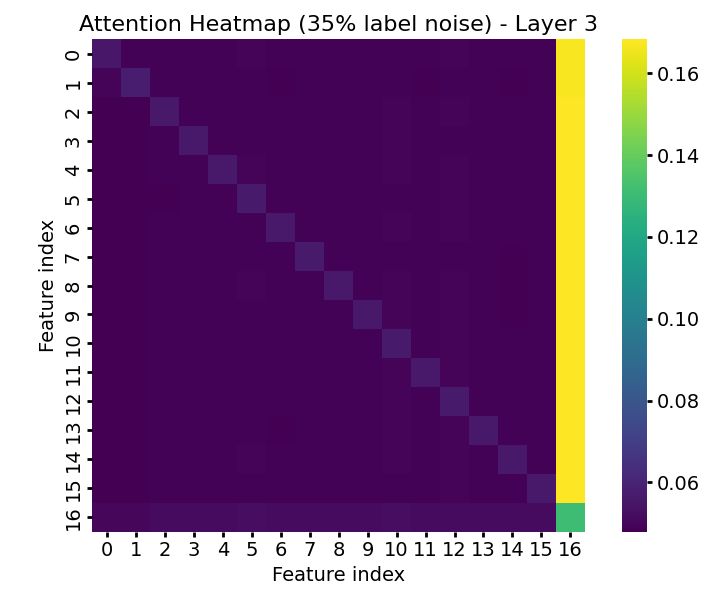}
\includegraphics[width=0.48\textwidth]{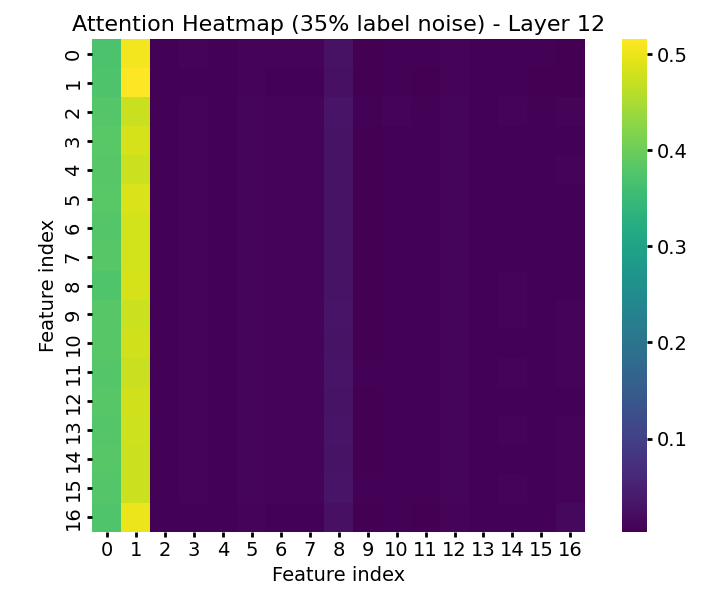}
\caption{Feature-wise attention weights heatmaps in layer \{3, 12\} of TabPFN when 35\% of targets are mislabeled. In each plot, the first 2 indices represent informative features and last index is the target label. Left: in layer 3; right: in layer 12.}
\label{fig:bylabelnoise_attn_heatmap}
\end{figure}

One interesting observation appears in the attention heatmaps that may help explain TabPFN's sharpness under the presence of label noise. As shown in Fig.~\ref{fig:bylabelnoise_attn_heatmap} when 35\% of labels are flipped, the target token attends less to itself (i.e. the cell of the last row and column) compared to attention received by other features in layer 3. In layer 12, the first informative feature (feature 0) receives less attention which may be due to this feature having stronger predictive power that leads to mislabeling. These imply that TabPFN is more uncertain towards the label and feature tokens when they contradict the patterns established by the feature-wise interactions. In such a case, TabPFN still learns to distribute most of its attention on 2 informative features eventually and it may rely more on consistent global context structure than on potentially corrupted tokens at some stage, effectively downweighting these unreliable tokens.

\section{Discussion}
\subsection{Why Does Robustness Persist Across the Parametric Tests?}
The main observation in this paper is that stable ROC-AUC is accompanied by consistent internal signals: across the parametric tests, attention remains structured and sharp rather than uniform and uncertain, and informative features remain highly ranked by the attention-based metrics. In the baseline analysis, this stability is further reflected in the qualitative diagnostics, where attention shifts from early emphasis on the label token toward later concentration on informative features, while feature-token embeddings become increasingly organized and eventually separate informative from random features. These observations suggest that robustness is not only an output-level phenomenon but is tied to a persistent internal preference for task-relevant structure.

A natural question, however, is why the model acquires this preference in the first place. One plausible interpretation is that the answer lies partly in how TabPFN is pre-trained. The TabPFN framework is trained on a very large collection of synthetic tabular datasets generated from a prior, and this prior is based on structural causal models (SCMs) mixed with Bayesian Neural Networks (BNNs), together with additional mechanisms for its causal generative process such as computational edge mapping, missing-value generation, and post-processing transformations \cite{tabpfn_v2}. The stated purpose of this design is to expose the model to many of the challenges that arise in real tabular data, including missing values, outliers, and uninformative attributes.
Therefore, the pre-training distribution is not a generic source of synthetic supervision but a carefully designed family of data-generating processes intended to encode structure into the tasks the model must solve.

Beyond the pre-training data distribution itself, the architecture of TabPFN as a Prior-Data Fitted Network can help make the model robust. A core objective of a PFN is to minimize losses across widely different datasets. If the model attempts to memorize a random feature's relationship to the target in one dataset, that exact relationship will be meaningless in a different dataset and can potentially increase the loss. Hence, to optimize the loss across all datasets, TabPFN learns to look for consistent relationships across the samples.
From these perspectives, one reasonable hypothesis is that pre-training on SCM-based synthetic tasks may encourage TabPFN to develop internal representations that preserve aspects of the underlying causal or structural relationships among variables \cite{swelam2025does}, while biasing prediction toward features that are consistently informative for the target under the generative mechanisms in the prior and away from variables that are disconnected or non-informative.
This would fit naturally with our empirical observations that random features receive very little meaningful attention even when the table becomes much wider.
Under this interpretation, TabPFN may be learning a powerful inductive bias: features that repeatedly participate in predictive relationships under the synthetic prior should attract attention, while features that do not should be ignored.

Ye et al. \cite{closerlook_tabpfnv2} offers a complementary insight: it argues that TabPFN can infer attribute relationships even when randomized attribute tokens are used at inference time, meaning that part of the model’s success may come from an ability to internalize relationship discovery during ICL itself.
In addition, TabPFN processes tabular inputs as an unordered set of features tokens rather than a fixed vector. This enforces a feature-permutation invariance so that no spurious correlations can be developed based on feature order or names.
At the same time, these explanations remain speculative.
We think the model’s robustness may arise not from one single mechanism, but from the interaction of a highly structured synthetic pre-training prior and a transformer architecture that is able to infer and refine inter-feature relationships during the forward pass.

This is the reason we think the ``why'' is still an open question even though we have observed the ``how''. The current evidence supports a meaningful working interpretation: TabPFN learns an internal preference for structurally useful features, and that preference may be shaped by SCM-based pre-training and then expressed through attention during inference. 
Without knowing the exact pre-training data or controlled re-training experiments under alternative priors, it is difficult to identify how much of the observed robustness comes specifically from the causal structure of the synthetic data, how much comes from the broader diversity of the prior, and how much comes from the transformer’s ICL mechanism. In other words, it is only reasonable to propose this as a mechanistic hypothesis based on the observations from our tests.

\subsection{Correlated Features as a Competitive Attribute}
The test with increasing number of nonlinearly correlated features reveals a qualitatively different kind of robustness challenge from the test with increasing number of random features. Random features are relatively easy to ignore because they carry no systematic relationship to the target. Correlated features, by contrast, share signal with the informative features and can therefore act as plausible substitutes. In our experiments, we purposely used only nonlinearly correlated features, which are harder to distinguish from plain informative features than linearly correlated ones are. It is visible in our attention-based metrics results: while random features remain weakly attended, correlated features can occasionally attract more attention than the least-attended informative feature, even though overall predictive performance remains stable.
This attention pattern may suggest a more flexible behavior of TabPFN while remaining robust: the model appears able to redistribute attention among multiple predictive candidates when the dataset contains redundant information. This internal competition among informative features and correlated features does not necessarily imply confusion to TabPFN's robust performance. One interpretation is that correlated features provide alternate routes to the same underlying predictive signal, similar to XGBoost's behavior when two features offer almost identical gain \cite{xgboost}.

However, this complicates the interpretation of the model. Such a challenge would be overlooked if we looked only at predictive performance such as ROC-AUC. The correlated-feature condition would appear almost as benign as the irrelevant-feature condition, but the attention metrics reveal a more nuanced reality: performance is stable, but the path by which the model arrives there is changing. As a result, it can be difficult to identify the true causal or explanatory variables. 

\subsection{Implications for Applied Modeling and Model Interpretability}
The results shown in this paper suggest that TabPFN can tolerate substantial feature clutter and moderate label corruption, without obvious degradation in ROC-AUC. This matters because messy tabular datasets are common in the real world, and many industrial problems fall into the class of small to medium scale tabular tasks for which TabPFN was originally designed.
Much of the literature on robustness in tabular learning evaluates models through predictive metrics under benchmarks. Those evaluations are useful, but they do not always reveal whether a model remains faithful to informative structure internally. By combining ROC-AUC with attention concentration, feature ranking, embedding projections, and SHAP visualizations, our study shows how robustness analysis can be made more interpretable. This is especially relevant for TFMs because they replace repeated fitting and tuning with inference time learning. Therefore, it is beneficial to understand whether the model is focusing on the correct and important information when data quality changes.

\subsection{Limitations and Future Work}
It is worth noting that in our parametric tests, the scale of the data does not reach the upper end of TabPFN v2's limit (up to 10000 rows and 500 features). This is due to the use of full attention inside the transformer, which is required to extract attention weight matrices for the core analysis in this paper, instead of a flash attention utilized in the original model \cite{tabpfn_v2, tabpfn_v2_5, dao2022flashattention}. Running the model is thus constrained by the practical memory cost of extracting full attention matrices such that the more challenging regime is not fully explored in our tests: simultaneously approaching upper limit settings in both dimensions where the model may plausibly struggle. Future work should test this joint regime, given more available hardware capacity.

In addition, all tests here use synthetic datasets to isolate factors (irrelevant features, correlation, sample size, and label noise) in a controlled manner. While this facilitates mechanistic interpretation, it remains important to evaluate whether the same internal robustness signatures appear on real-world datasets and under practical ``stress'' transformations (i.e., if we have a base dataset, how does the performance change if the model is used to predict on a stressed dataset). Extending the analysis to real-world settings would help clarify how best to use TabPFN in practice for stable and accurate predictions, and thus to better understand how one can benefit from TabPFN’s robustness.

Lastly, the limitation returns to the main ``why'' question. As discussed above, we can propose a plausible explanation rooted in SCM-based pre-training and ICL, but we cannot definitively isolate which part of the training design is responsible for the robustness patterns we observe. A more decisive test of the causal pre-training hypothesis would likely require pre-training comparable models under alternative synthetic data generation schemes and then comparing their internal robustness signatures. Those are beyond the scope of the current work, but can move the field from plausible interpretation to stronger explanation.

\section{Conclusion}
This paper asked practically important questions: can TabPFN handle noisy tabular data and if so, how? To address them, we conducted a robustness study under controlled synthetic parametric tests that mirror several common data quality issues in the real world, including increasing feature dimensionality through irrelevant and correlated predictors, varying sample size, and increasing the proportion of mislabeled targets. Across all tests, predictive performance remains stable at high levels, while attention-based metrics show that the model continues to prioritize informative features and maintains structured, non-uniform attention.

The findings in this paper complement the broader motivation for tabular foundation models: shifting from repeated per-dataset fitting toward inference-time adaptation via attention and ICL.
The value of these findings lies not only in the preservation of performance metrics such as ROC-AUC, but also in the mechanistic evidence accompanying that stability. The baseline analyses show that attention gradually concentrates on useful features across layers, feature-token embeddings become more separable, and SHAP values align with the same informative dimensions. The parametric tests then show that this internal structure persists under several forms of variation.
At the same time, the study raises a deeper question about mechanism. A plausible interpretation is that the model’s robustness is related to its pre-training on a synthetic prior built from structural causal models, together with its ability to infer inter-feature relationships during ICL. 
In this sense, the paper contributes both a robustness result and a direction to explain how and why that robustness may arise.

\section*{Acknowledgments}
We would like to thank Travis Ens for his continued involvement throughout this project, from its early development to the revision of this manuscript, and for his valuable feedback and constructive suggestions, which helped improve the organization and presentation of this paper.

\newpage
\bibliographystyle{plain}
\bibliography{refs}

\newpage
\appendix
This appendix contains additional documentation and results that detail and support the findings presented in the main paper. The following sections include:
\begin{itemize}
 \item Appendix A: Detailed implementation descriptions for our experiments.
 \item Appendix B: Additional test results and analysis using the same model with different data generation process.
\end{itemize}
\section{Experimental Details}
\paragraph{Synthetic Data Generation.} All tests in the main paper were conducted using synthetic datasets with controllable settings. To achieve this, we use the $\texttt{make\_\allowbreak classification}$ function$\footnote{See the official scikit-learn documentation for \texttt{make\_classification}: \url{https://scikit-learn.org/stable/modules/generated/sklearn.datasets.make_classification.html}}$ from scikit-learn to generate the datasets. Section 3 discusses the settings of tests and here we provide more details about the construction of nonlinearly correlated features and labels used in the main paper. Although the $\texttt{make\_classification}$ function natively supports creation of linearly correlated features, we implemented a simple nonlinear redundancy generator to increase the complexity of the datasets to simulate real-world situations where data features are often not linearly dependent. Correlated features were derived from an informative feature or both informative features using one of four nonlinear transformations randomly:
\begin{itemize}
 \item Square: $x_{corr} = x^2_{infor}$
 \item Interaction: $x_{corr} = x_{infor,1} \times x_{infor, 2}$
 \item Sine: $x_{corr} = \sin(x_{infor})$
 \item Exponential: $x_{corr} = \exp(\text{clip}(x_{infor}, -3, 3))$.
\end{itemize}
Each correlated feature was then scaled by a random factor $\in [0.5, 2.0]$ and shifted by a random bias based on its standard deviation to ensure each correlated feature is different and cannot be trivially mapped back to the informative features.

For label construction, the $\texttt{make\_classification}$ function generates the informative features by forming Gaussian clusters in the informative subspace. Here, a Gaussian cluster refers to a group of samples concentrated around a common center, with informative feature values drawn from a normal distribution. In our balanced binary setting, we use one cluster per class by setting \texttt{n\_clusters\_per\_class}=1, so each class is represented by a single Gaussian-distributed group of samples in the informative feature space. Unless otherwise stated, we also use the default label-flip rate \texttt{flip\_y}=0.01, meaning that 1\% of labels are randomly assigned.

\paragraph{Model Configuration Remark.} Although the most current TabPFN models are v2.5 models \cite{tabpfn_v2_5} where feature group size = 3 are used, to facilitate these tests with emphasis and visualizations on attention on and from each feature, we use a TabPFN v2 checkpoint with feature group size = 1, similar to the approach in \cite{closerlook_tabpfnv2}. Furthermore, to eliminate any artifacts from ensembling and randomness, we use 1 estimator and disable feature shuffling when fitting the model. Model parameter $\texttt{ignore\_pretraining\_limit}$ is set to $\texttt{True}$ to bypass TabPFN v2's limit on the number of features of 500 for test cases with 512 total number of features. Although the official document for TabPFN cautions that TabPFN may not perform well on data outside the pre-training range, the model's limit is not strict. Strong results were reported on benchmarks that exceed the model's limit in \cite{tabpfn_v2_5}.

\paragraph{Compute Resources.} All computational tests reported in this paper were conducted on Azure Machine Learning platform using the Standard\_NC40ads\_H100\_v5 instance.

\section{Additional Parametric Tests}
In this section, we provide further test results to evaluate the robustness of TabPFN under a range of configurations to generate synthetic datasets using $\texttt{make\_classification}$ function. To investigate if similar conclusions can be drawn under different scenarios, three additional settings for the data generation process are considered here:
\begin{enumerate}
 \item Using a different random seed. This changes the random realization of the dataset, varying the relationship between informative features and class labels. 
 \item Increasing $\texttt{n\_clusters\_per\_class}$ from 1 to 2. This introduces multimodality within each class, making the class structure more complex and the decision boundary more nonlinear.
 \item Decreasing $\texttt{class\_sep}$ from 1 to 0.5. This reduces the separation between classes, weakening the relationship between informative features and class labels. 
\end{enumerate}
The results of ROC AUC and attention ratio of informative features over other features in layer 12 based on these data generation processes are summarized in the tables below for three different parametric tests described in the main paper. All tables show stable metrics and consistent results as seen in the main paper so that TabPFN is robust to datasets generated with different logic. For example, Table.~\ref{tab:appendix_rand_1} and Table.~\ref{tab:appendix_rand_2} indicates predictive performance ROC AUC does not drop and attention ratio stays large as number of random features increases for all 3 data generation processes. Similar observations are presented for other parametric tests. We note that for the data generation process where class separation value is decreased, lower ROC AUC values are shown. It is due to the fact that these datasets are harder to predict by design since the relationship between informative features and class labels are weakened intentionally. However, ROC AUC values are stable across the parametrized values.
 
\begin{table}[h]
\centering
\begin{tabular}{l | c c c c c c c}
\hline
\multicolumn{8}{c}{ROC AUC} \\
\hline
Data generation process & \multicolumn{7}{c}{Total number of features} \\
\cline{2-8}
                        & 8 & 16 & 32 & 64 & 128 & 256 & 512 \\
\hline
Different seed &0.995 &0.992 &0.996 &0.990 &0.996 &0.997 &0.997\\
Larger number of clusters &0.986 &0.985 &0.981 &0.987 &0.978 &0.985 &0.984\\
Smaller class separation &0.888 &0.890 &0.882 &0.886 &0.886 &0.874 &0.875\\
\hline
\end{tabular}
\caption{ROC AUC using different data generation processes and various total numbers of features.}
\label{tab:appendix_rand_1}
\end{table} 

\begin{table}[h]
\centering
\begin{tabular}{l | c c c c c c c}
\hline
\multicolumn{8}{c}{Attention Ratio of Informative over Random} \\
\hline
Data generation process & \multicolumn{7}{c}{Total number of features} \\
\cline{2-8}
                        & 8 & 16 & 32 & 64 & 128 & 256 & 512 \\
\hline
Different seed  &47.997 &41.788 &159.154 &178.965 &264.194 &227.691 &167.835\\
Larger number of clusters &61.505 &131.740 &208.987 &192.664 &317.481 &191.223 &126.268\\
Smaller class separation &65.657 &68.450 &152.730 &122.236 &180.921 &122.194 &141.558\\
\hline
\end{tabular}
\caption{Attention ratio of informative features over random features in layer 12 using different data generation processes and various total numbers of features.}
\label{tab:appendix_rand_2}
\end{table} 

\begin{table}[h]
\centering
\begin{tabular}{l | c c c c c c c}
\hline
\multicolumn{8}{c}{ROC AUC} \\
\hline
Data generation process & \multicolumn{7}{c}{Number of correlated features} \\
\cline{2-8}
                        & 2 & 4 & 8 & 16 & 32 & 64 & 128 \\
\hline
Different seed &0.997 &0.997 &0.997 &0.997 &0.997 &0.997 &0.997\\
Larger number of clusters &0.985 &0.986 &0.986 &0.985 &0.985 &0.985 &0.984\\
Smaller class separation &0.875 &0.874 &0.875 &0.876 &0.875 &0.873 &0.873\\
\hline
\end{tabular}
\caption{ROC AUC using different data generation processes and various numbers of correlated features.}
\label{tab:appendix_corr_1}
\end{table} 

\begin{table}[h]
\centering
\begin{tabular}{l | c c c c c c c}
\hline
\multicolumn{8}{c}{Attention Ratio of Informative over Correlated+Random} \\
\hline
Data generation process & \multicolumn{7}{c}{Number of correlated features} \\
\cline{2-8}
                        & 2 & 4 & 8 & 16 & 32 & 64 & 128 \\
\hline
Different seed &81.124 &67.135 &63.581 &50.756 &19.162 &15.262 &6.182\\
Larger number of clusters &73.133 &74.398 &55.768 &44.370 &32.758 &26.490 &21.276\\
Smaller class separation &150.600 &67.238 &51.935 &23.602 &13.451 &12.371 &11.122\\
\hline
\end{tabular}
\caption{Attention ratio of informative features over correlated and random features in layer 12 using different data generation processes and various numbers of correlated features.}
\label{tab:appendix_corr_2}
\end{table} 
 
\begin{table}[h]
\centering
\begin{tabular}{l | c c c c c c c}
\hline
\multicolumn{8}{c}{ROC AUC} \\
\hline
Data generation process & \multicolumn{7}{c}{Total number of training rows} \\
\cline{2-8}
                        & 3000 & 4500 & 6000 & 7500 & 9000 & 10500 & 12000 \\
\hline
Different seed &0.993 &0.996 &0.994 &0.983 &0.995 &0.994 &0.994\\
Larger number of clusters &0.984 &0.983 &0.972 &0.984 &0.983 &0.991 &0.982\\
Smaller class separation &0.978 &0.995 &0.928 &0.991 &0.926 &0.873 &0.968\\
\hline
\end{tabular}
\caption{ROC AUC using different data generation processes and various total number of training rows.}
\label{tab:appendix_row_1}
\end{table} 

\begin{table}[h]
\centering
\begin{tabular}{l | c c c c c c c}
\hline
\multicolumn{8}{c}{Attention Ratio of Informative over Random} \\
\hline
Data generation process & \multicolumn{7}{c}{Total number of training rows} \\
\cline{2-8}
                        & 3000 & 4500 & 6000 & 7500 & 9000 & 10500 & 12000 \\
\hline
Different seed &60.895 &53.664 &48.727 &78.906 &48.727 &34.798 &86.640\\
Larger number of clusters &60.354 &93.889 &90.220 &79.762 &88.258 &92.712 &119.712\\
Smaller class separation &91.052 &88.664 &118.795 &123.444 &134.075 &126.989 &108.147\\
\hline
\end{tabular}
\caption{Attention ratio of informative features over random features in layer 12 using different data generation processes and various total numbers of training rows.}
\label{tab:appendix_row_2}
\end{table} 

\begin{table}[h]
\centering
\begin{tabular}{l | c c c c c c c}
\hline
\multicolumn{8}{c}{ROC AUC} \\
\hline
Data generation process & \multicolumn{7}{c}{Fraction of label noise} \\
\cline{2-8}
                        & 0.05 & 0.1 & 0.15 & 0.2 & 0.25 & 0.3 & 0.35 \\
\hline
Different seed &0.999 &0.999 &0.999 &0.999 &0.999 &0.999 &0.999\\
Larger number of clusters &0.978 &0.977 &0.974 &0.977 &0.972 &0.975 &0.967\\
Smaller class separation &0.931 &0.931 &0.931 &0.931 &0.930 &0.931 &0.924\\
\hline
\end{tabular}
\caption{ROC AUC using different data generation processes at various fractions of label noise.}
\label{tab:appendix_label_1}
\end{table} 

\begin{table}[h]
\centering
\begin{tabular}{l | c c c c c c c}
\hline
\multicolumn{8}{c}{Attention Ratio of Informative over Random} \\
\hline
Data generation process & \multicolumn{7}{c}{Fraction of label noise} \\
\cline{2-8}
                        & 0.05 & 0.1 & 0.15 & 0.2 & 0.25 & 0.3 & 0.35 \\
\hline
Different seed &76.172 &80.865 &84.116 &60.062 &78.018 &60.809 &61.788\\
Larger number of clusters &61.319 &45.238 &48.377 &29.345 &32.015 &60.531 &45.640\\
Smaller class separation &109.531 &111.949 &110.681 &83.177 &76.726 &56.233 &57.198\\
\hline
\end{tabular}
\caption{Attention ratio of informative features over random features in layer 12 using different data generation processes at various fractions of label noise.}
\label{tab:appendix_label_2}
\end{table}

\end{document}